\documentclass[10pt,twocolumn,letterpaper]{article}

\usepackage{cvpr}              %

\usepackage{comment}
\usepackage{arydshln}
\usepackage{graphicx}
\usepackage{amsmath}
\usepackage{amssymb}
\usepackage{caption}
\usepackage{multirow}
\usepackage{colortbl} %
\usepackage{xcolor}   %

\newcommand{\method}{\textsc{DualVision}\xspace}
\newcommand{\dataset}{\textsc{DV-204K}\xspace}
\newcommand{\benchmark}{\textsc{DV-500}\xspace}

\definecolor{cvprblue}{rgb}{0.21,0.49,0.74}
\usepackage[pagebackref,breaklinks,colorlinks,allcolors=cvprblue]{hyperref}

\def\paperID{33621} %
\def\confName{CVPR}
\def\confYear{2026}

\title{\method: RGB-Infrared Multimodal Large Language Models for Robust Visual Reasoning}

\author{Abrar Majeedi\textsuperscript{1}, 
Zhiyuan Ruan\textsuperscript{2}, Ziyi Zhao\textsuperscript{2}, Hongcheng Wang\textsuperscript{2}, Jianglin Lu\textsuperscript{3}, Yin Li\textsuperscript{1} \\
{\textsuperscript{1}University of Wisconsin-Madison, \
\textsuperscript{2}Amazon, \
\textsuperscript{3}Northeastern University}
}

\begin{document}

\twocolumn[{%
\renewcommand\twocolumn[1][]{#1}%
\maketitle
\vspace{-2em}
\begin{center}
\includegraphics[width=0.8\linewidth]{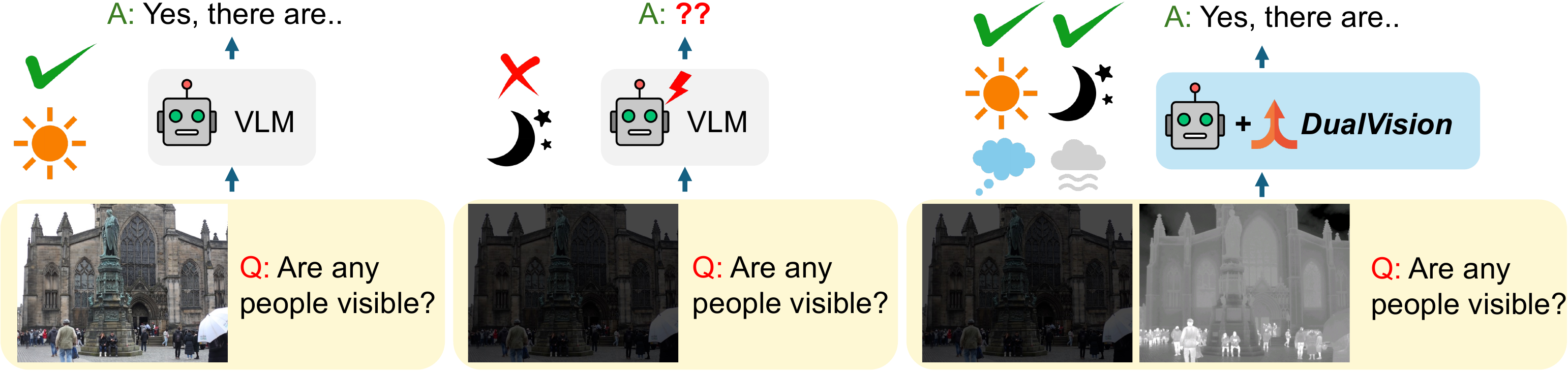}\vspace{-0.5em}
\captionof{figure}{When visibility degrades in RGB images (\eg, at night), MLLMs struggle to ``see and reason'', limiting their reliability in many real-world applications such as autonomous driving. By complementing RGB with infrared data, 
\method enables robust visual perception and reasoning while reducing computation by $\sim$75\% compared to na\"ive fusion.  %
}
\label{fig:teaser_figure}
\end{center}%
}]

\begin{abstract}
Multimodal large language models (MLLMs) have achieved impressive performance on visual perception and reasoning tasks with RGB imagery, yet they remain fragile under common degradations, such as fog, blur, or low-light conditions. Infrared (IR) imaging, a well-established complement to RGB, offers inherent robustness in these conditions, but its integration into MLLMs remains underexplored. 
To bridge this gap, we propose \method, a lightweight fusion module that efficiently incorporates IR--RGB information into MLLMs via patch-level localized cross-attention.
To support training and evaluation and to facilitate future research, we also introduce \dataset, a dataset of $\sim$25K publicly available aligned IR--RGB image pairs with $204K$ modality-specific QA annotations, and \benchmark, a benchmark of $500$ IR--RGB image pairs with 500 QA pairs designed for evaluating cross-modal reasoning.
Leveraging these datasets, we benchmark both open- and closed-source MLLMs and demonstrate that \method delivers strong empirical performance under a wide range of visual degradations.
Our code and dataset are available at \href{https://abrarmajeedi.github.io/dualvision/}{https://abrarmajeedi.github.io/dualvision}.
\end{abstract}

\section{Introduction}
\label{sec:intro}
Multimodal large language models (MLLMs)~\cite{yin2024survey} represent an increasingly important class of vision language models (VLMs) that connect visual and textual data through large language models (LLMs). MLLMs have achieved strong performance across many visual recognition tasks such as object recognition, visual grounding, and visual question answering, enabling broad applications in robotics~\cite{kim2025openvla}, autonomous driving~\cite{cui2024survey}, and digital health~\cite{alsaad2024multimodal}. Most existing MLLMs, however, rely exclusively on RGB imagery as their visual input, drawing from large-scale web datasets composed primarily of well-lit, natural scenes. 

Although RGB imagery provides rich color and texture information and has driven impressive generalization in standard benchmarks, it exposes a critical weakness: their reliability drops sharply when inputs are degraded by adverse visual conditions~\cite{usama_analysing_2025}. This vulnerability stems from RGB's dependence on visible light and its susceptibility to optical distortions. Common examples include low-light environments, motion- or defocus-induced blur, and non-ideal weather such as rain or fog. 
These degradations are not rare anomalies but frequent realities in practical deployment, particularly in domains such as transportation, surveillance, and health, where robustness is paramount. For example, autonomous vehicles must sustain robust perception at night and in adverse weather, while home-based health monitoring systems must function effectively under poor lighting and motion blur.

Infrared (IR) imaging offers a valuable complement: by capturing electromagnetic radiation beyond the visible spectrum, IR can remain effective in darkness, fog, and other challenging environments faced by RGB imagery~\cite{shin_sparse_2019, grabner_wavelength_2011}. However, IR imagery lacks the fine-grained appearance details and semantic richness that RGB captures under favorable conditions.
Fusing RGB and IR signals thus provides a promising pathway towards more robust visual perception and reasoning by leveraging their complementary strengths. This fusion has been widely explored in traditional vision tasks such as recognition, detection, and segmentation~\cite{wu2017rgb,zhang_object_2020,wang_sgfnet_2023}, and as a strategy for mitigating degradations through complementary sensing~\cite{tang2015high,ramanagopal2024theory}.
However, its integration into MLLMs, particularly to overcome the limitations of RGB under \textit{degraded visual conditions}, remains largely underexplored to date.

Developing MLLMs that can jointly comprehend RGB and IR data faces three key challenges. \textit{First}, there is no principled design for a fusion mechanism that maintains spatial alignment between RGB and IR modalities while adaptively prioritizing the informative signals for MLLMs. \textit{Second}, progress in IR-RGB perception is hindered by the scarcity of large-scale, semantically rich datasets. Existing datasets, often developed for detection or segmentation, tend to be narrow in scope, lack linguistic annotations, and are not aligned with the instruction-tuning paradigm that drives recent advances in MLLMs. \textit{Third}, the absence of standardized IR-RGB benchmarks for vision language tasks leads to inconsistent evaluation, particularly under visual degradations, making it difficult to rigorously assess robustness across varying visual conditions.

In this paper, we address key challenges in developing IR-RGB MLLMs for robust visual reasoning. As illustrated in Fig. \ref{fig:teaser_figure}, we propose \emph{\method}, a lightweight fusion approach that leverages multi-scale localized cross-attention to selectively route information between aligned IR and RGB tokens. Our design exploits the inherent spatial structure of visual data by employing progressively expanding local attention radii across different fusion levels. This hierarchical approach enables precise local correspondence at fine scales while capturing broader contextual relationships at coarser ones. The result is a unified IR-RGB representation that avoids the quadratic overhead of na\"ive concatenation, while the localized cross-attention mechanism provides inductive biases that strengthen cross-modal alignment, particularly under degraded RGB conditions.

Complementing our approach, we introduce \emph{\dataset} and \emph{\benchmark}, a new dataset suite designed to facilitate both instruction tuning and systematic evaluation of MLLM's robustness under visual degradations. Our datasets provide diverse, well-aligned IR-RGB image pairs with modality-aware question-answer pairs, allowing the study of both general reasoning and degradation-specific performance.

\smallskip
\textbf{Our contributions} are summarized as follows.
\begin{enumerate}
    \item Our work is among the first to develop MLLMs that integrate RGB and IR modalities for robust visual reasoning under visual degradations (\eg, blur, low-light, and fog). 
    \item We introduce \method, a lightweight IR-RGB fusion module with enhanced robustness to degradations, while remaining compatible with existing MLLMs.
    \item To support training and evaluation of IR-RGB MLLMs, we create and release two datasets: 1) \dataset: A dataset of $\sim$25K aligned IR–RGB image pairs with $\sim$204K modality-aware question-answer annotations designed for instruction tuning and 2) \benchmark: A carefully curated evaluation benchmark featuring 500 IR-RGB image pairs with 500 associated QA pairs.
    \item Leveraging our datasets and through extensive experiments, 
    we demonstrate strong empirical results of \method under various degradations.
\end{enumerate}

\section{Related Work}
\label{sec:related_work}

\noindent \textbf{VLMs and MLLMs.} Modern VLMs align visual and textual modalities through contrastive and generative pretraining. CLIP~\cite{radford_learning_2021} pioneers large-scale contrastive learning, enabling open-vocabulary recognition. Recent efforts, including ImageBind~\cite{girdhar2023imagebind}, LanguageBind~\cite{zhu_languagebind_2023} extend this paradigm to additional modalities including IR, depth and audio. However, these works focus on broad modality binding~\cite{girdhar2023imagebind, zhu_languagebind_2023}, rather than more complex reasoning.

More recent MLLMs such as LLaVA~\cite{liu_visual_2023}, BLIP-2~\cite{li_blip-2_2023}, and Flamingo~\cite{alayrac_flamingo_2022} integrate vision encoders with LLMs to support open-ended visual reasoning. Instruction tuning further improved alignment with human intent. InstructBLIP~\cite{dai_instructblip_2023}, MiniGPT-4~\cite{zhu_minigpt-4_2023}, and LLaVA-1.5~\cite{liu_improved_2024} demonstrated that curated instruction datasets enhance reasoning and zero-shot generalization. While most MLLMs rely solely on RGB imagery, a few very recent works, including IR-LLaVA~\cite{jiang_infrared-llava_2024} and IRGPT~\cite{cao_irgpt_2025}, have begun exploring IR-based MLLMs. However, these approaches operate on IR-only inputs and discard RGB, overlooking the complementary sensing capabilities of the two modalities. In contrast, we aim to develop IR–RGB MLLMs that fully exploit cross-modal synergy for robust visual reasoning.

\begin{figure*}[th!]
    \centering
    \includegraphics[width=0.86\linewidth]{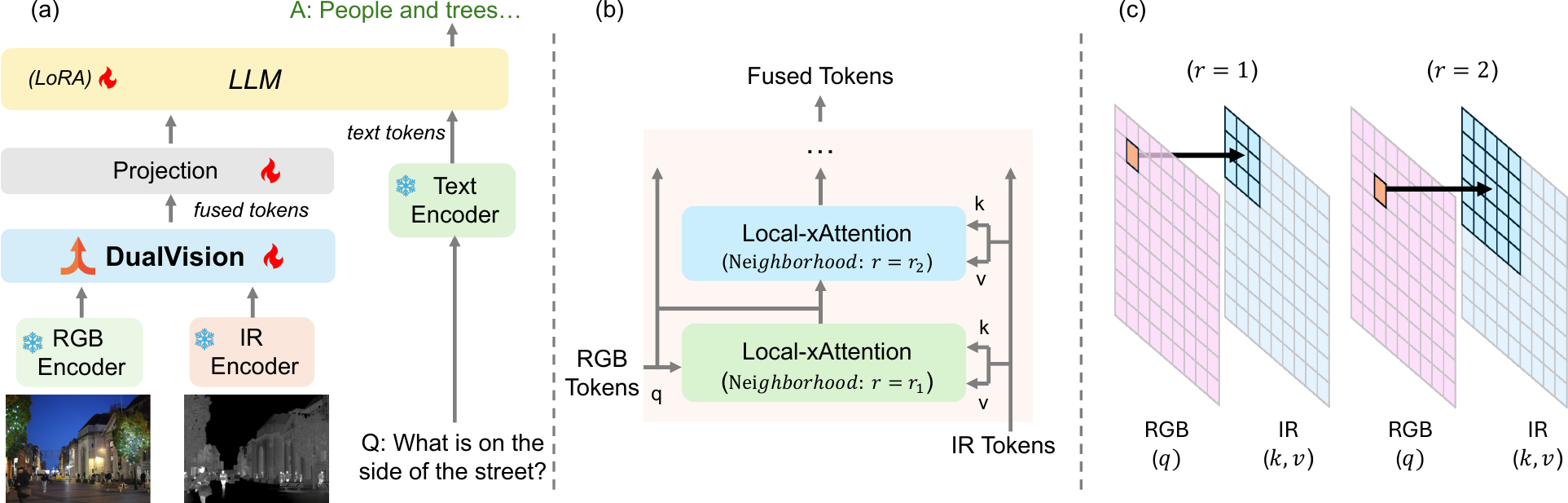}\vspace{-0.5em}
    \caption{Overview of \textbf{\method}. \textbf{(a)} shows how \method integrates into a MLLM to fuse RGB and IR image tokens for robust visual reasoning. \textbf{(b)} illustrates the multi-scale localized cross-attention module, where RGB tokens serve as queries and IR tokens as keys and values. \textbf{(c)} visualizes the spatially aligned RGB–IR token grids over which localized cross-attention is performed. %
    }\vspace{-1.5em}
    \label{fig:main_fig}
\end{figure*}

\smallskip
\noindent \textbf{Multimodal Fusion in MLLMs.} Extending vision models beyond RGB requires effective fusion of complementary modalities. To integrate multiple modalities, recent MLLMs harness the flexibility of generic architecture, \eg, Transformer~\cite{vaswani_attention_2017}. In these models, signals from each modality are first tokenized using modality-specific encoders; the resulting tokens are then interleaved and passed to the Transformer, where the self-attention mechanism performs multimodal fusion~\cite{nagrani_attention_2021}.

However, self-attention scales quadratically with token count~\cite{vaswani_attention_2017}, thus rendering vanilla concatenation based multimodal fusion computationally expensive. Efficient variants such as local or sparse attention reduce cost but often weaken cross-modal alignment. Seminal works, such as Swin Transformer~\cite{liu_swin_2021} mitigate this via windowed self-attention, though it remains intra-modal. Extensions like SwinFusion~\cite{ma_swinfusion_2022} achieve IR–RGB fusion for reconstruction tasks but are heavy and limited to low-level vision. More recent multimodal systems explore lightweight and flexible fusion designs. PandaGPT~\cite{su_pandagpt_2023} performs element-wise addition of visual and textual tokens for compact fusion, while our prior work~\cite{liu_pave_2025} focuses on video-LLMs and employs cross-attention blocks to incorporate additional modalities (\eg, audio or depth) without retraining the base architecture, enabling scalable multimodal integration.

\smallskip
\noindent \textbf{Robustness to Visual Corruptions.} %
The vulnerability of vision systems to common image corruptions has been systematically documented by~\cite{hendrycks_benchmarking_2018}, who introduced the ImageNet-C benchmark with 19 corruption types across four categories: noise, blur, weather, and digital distortions. Usama et al. \cite{usama_analysing_2025} extended this analysis to VLMs, revealing vulnerability patterns across different tasks. 
To address such vulnerabilities, prior work has largely pursued two directions: degradation-aware processing and input restoration. Quality-aware networks~\cite{wang_cqa-face_2022} adapt feature extraction to estimate corruption levels and leverage that information during inference, while restoration-based methods~\cite{sun_rethinking_2022} attempt to reconstruct clean inputs prior to inference. While effective for single-modality vision, these approaches add inference cost and fail to exploit complementary sensing.

\smallskip
\noindent \textbf{IR-RGB for Robust Perception.} 
Cross-modal sensing offers a promising approach to enhance robustness by combining complementary modalities.
IR-RGB fusion has proven effective for object detection~\cite{wang_mixture_2025} and semantic segmentation~\cite{wang_sgfnet_2023}, especially in autonomous driving~\cite{brenner_rgb-d_2023}, where thermal imaging compensates for RGB limitations in low-light conditions. Yet, this success has not translated to MLLMs, where the difficulty lies not only in perception but also in cross-modal reasoning and grounding.

A key barrier is the absence of suitable datasets. Existing vision-language benchmarks rely solely on RGB data, while robustness datasets~\cite{hendrycks_benchmarking_2018,usama_analysing_2025} test corruption within a single modality. Although IR-RGB datasets exist for detection and segmentation~\cite{jia_llvip_2021,shivakumar_pst900_2020}, there is an unmet need for IR-RGB datasets with question-answer pairs necessary for multimodal reasoning evaluation.

To bridge the gaps, we propose \emph{\method}, an effective and efficient IR-RGB fusion module designed for MLLMs. We also create new datasets for training and evaluating IR-RGB reasoning under challenging conditions.

\section{Design of \method}

\method, as shown in Fig.\ \ref{fig:main_fig}, presents a lightweight RGB-IR fusion module designed for MLLMs. Instead of interleaving tokens from IR and RGB, \method performs \emph{multi-scale localized cross-attention}, allowing each RGB patch token to attend only to spatially corresponding IR regions. \method injects complementary IR cues where they are relevant, has low computational overhead, and remains compatible with many existing MLLMs.

Let $X_{\text{RGB}} \in \mathbb{R}^{H \times W \times 3}$ and $X_{\text{IR}} \in \mathbb{R}^{H \times W \times 1}$ denote spatially aligned RGB and IR inputs.
A pre-trained vision encoder $E$ then maps each input to a sequence of patch tokens:
\begin{equation}
\small
Z^{\text{RGB}} = E(X_{\text{RGB}}), \quad
Z^{\text{IR}} = E(X_{\text{IR}}),
\end{equation}
with $Z^{\text{RGB}}, Z^{\text{IR}} \in \mathbb{R}^{N \times d}$. Each token corresponds to a patch feature that can be traced back to a specific location in the 2D image plane.
With these patch coordinates, we define a local neighborhood for each RGB token and performs \emph{local cross-attention} over the corresponding IR tokens.\footnote{A shared encoder implies implicit alignment, however, if separate encoders are employed, token grids can be aligned by interpolation.%
}
As shown in Fig.\ \ref{fig:main_fig}~(c), each RGB token attends only to IR tokens within a radius-$r$ region centered at its location, enforcing spatially aligned fusion. 
This design injects IR cues precisely where they are informative, while reducing the compute cost of global cross-modal interaction.

\smallskip
\noindent \textbf{2D Local Cross-Attention.}
\label{sec:local_xattn}
Let $\{z^{\text{RGB}}_u\}_{u=1}^N$ and $\{z^{\text{IR}}_v\}_{v=1}^N$ denote the RGB and IR token embeddings, 
and let $\{p^{\text{RGB}}_u\}$, $\{p^{\text{IR}}_v\} \subset \mathbb{Z}^2$ denote their patch coordinates. 
The neighborhood of an RGB token in the IR token grid is defined as
\begin{equation}
\small
    \mathcal{N}_r(u) = \left\{ v \in \{1, \dots, N\} : \rho\!\left(p^{\text{IR}}_v, \,\phi(p^{\text{RGB}}_u)\right) \le r \right\},
\end{equation}
where $\rho$ is the 2D Euclidean distance, and $\phi$ maps RGB coordinates to the IR grid if resolutions differ.

Leveraging the neighborhood $\mathcal{N}_r(u)$, each RGB token serves as the query while keys and values are drawn from the IR tokens in its restricted neighborhood. We then compute the projected query, key, and value as:
\begin{equation*}
\small
    q_u = z^{\text{RGB}}_u W_Q, 
    \ \ K_u = Z^{\text{IR}}_{\mathcal{N}_r(u)} W_K, 
    \ \ V_u = Z^{\text{IR}}_{\mathcal{N}_r(u)} W_V,
\end{equation*}
with $W_Q \in \mathbb{R}^{d \times d_k}$, $W_K \in \mathbb{R}^{d \times d_k}$, and $W_V \in \mathbb{R}^{d \times d_v}$.

Consequently, the attention weights are computed and applied to obtain the fused representations:
\begin{equation}
\small
\begin{aligned}
    &\alpha_u = \mathrm{softmax}\!\left( \frac{q_u K_u^\top}{\sqrt{d_k}} \right),\quad o_u = \alpha_u V_u, \\
    &\tilde{z}^{\text{RGB}}_u = z^{\text{RGB}}_u + o_u W_O
\end{aligned}\label{eq:xattn}
\end{equation}
where $W_O \in \mathbb{R}^{d_v \times d}$. Stacking the fused tokens over all RGB tokens yields the fused representation $Z_{\text{fused}} \in \mathbb{R}^{N \times d}$.

\smallskip
\noindent \textbf{Multi-Scale Cross-Attention.} 
To capture interactions across local regions with varying size while preserving locality, we employ multiple 2D local cross-attention blocks sequentially with progressively increased radii $r_1 \le \cdots \le r_L$. This is initialized as
$
Z^{(0)} = Z^{\text{RGB}},
$
and updated at each layer $\ell = 1, \dots, L$, through local cross-attention:
\begin{equation}
\small
\begin{aligned}
\hat Z^{(\ell)} &= Z^{(\ell-1)} 
    + \mathrm{xAttn}_{r_\ell}\!\left(\mathrm{LN}\!\left(Z^{(\ell-1)}\right),\, Z^{\mathrm{IR}}\right),\\
Z^{(\ell)} &= \hat Z^{(\ell)} 
    + \mathrm{FFN}\!\left(\mathrm{LN}\!\left(\hat Z^{(\ell)}\right)\right),
\end{aligned}
\end{equation}
where $\mathrm{xAttn}$ is the 2D local cross-attention in Eq. \ref{eq:xattn}, and $\mathrm{LN}$ and $\mathrm{FFN}$ denote LayerNorm and MLP, respectively. 
The final fused representation is given by $Z_{\text{fused}} = Z^{(L)}$.

This hierarchical design, together with skip connections, preserves the local spatial alignment while allowing the model to
have different receptive fields rather than being confined to a single arbitrarily fixed neighborhood. In practice, we rely on 
3 sequential blocks with $r \in [1, 2, 3 ]$.

\smallskip
\noindent \textbf{Interface with LLM}. To interface the fused IR-RGB tokens $Z_{L}$ with a LLM, we employ a linear projection $P$, trained jointly with the fusion module. During training, a LoRA adapter~\cite{hu_lora_2021} is attached to the LLM and only its parameters $\Theta_{\text{LoRA}}$ are updated. 
At inference time, fused tokens $Z_{L}$ are first projected and then concatenated with text tokens $Q$ from a text query $q$. The resulting multimodal sequence is then passed to the LLM, which autoregressively generates the answer sequence:
\begin{equation}
\small
    y_{1:T} \sim \text{LLM}(P(Z_{\text{fused}}), Q).
\end{equation}

\noindent \textbf{Training Objective.} Model training follows the standard next-token prediction loss over the answer sequence conditioned on both IR and RGB image and the text query $q$:
\begin{equation}
\small
    \mathcal{L}(\Theta) = - \Sigma_{t=1}^T \ \log p_\Theta \big(y_t \mid X_{\text{RGB}}, X_{\text{IR}}, q, y_{<t}\big),
\end{equation}
where $\Theta = \{\Theta_{\text{LoRA}}, P, \{W_Q, W_K, W_V, W_O\}_{1:L}\}
$.

\smallskip
\noindent \textbf{Data Augmentation}. To enhance robustness in degraded conditions, we adopt a degradation-aware training scheme. Specifically, with probability $p_d$ (=$0.25$ in our experiments), a degradation type $\delta \in \{\text{blur}, \text{darkness}, \text{fog}\}$ 
with severity $s \in \{\text{low}, \text{moderate}, \text{high}, \text{highest}\}$ 
is applied to the RGB image during training, \ie,
$
X'_{\text{RGB}} = T_{\delta, s}(X_{\text{RGB}})
$, which explicitly encourages the model to rely more on the IR features when RGB is unreliable.

\smallskip
\noindent \textbf{Computational Efficiency}.
MLLMs with Transformers~\cite{vaswani_attention_2017} have a quadratic cost to their input sequence length.
RGB-IR concatenation yields $2N$ tokens with cost $\mathcal{O}((2N)^2)=\mathcal{O}(4N^2)$. In contrast, \method fuses both modalities into $N$ tokens, reducing the cost to $\mathcal{O}(N^2)$, a 4$\times$ reduction. The fusion step uses local cross-attention with complexity $\mathcal{O}(N w^2)$ for window size $w<N$, which is negligible compared to the repeated LLM attention blocks. 
This computational advantage is reflected in practice: as shown in Table~\ref{tab:compute_comparison}, \method adds almost no parameters while substantially reducing TFLOPs, all while maintaining competitive performance. Implementation details are provided in the supplement.

\begin{table}[t!]
\centering
\resizebox{0.7\linewidth}{!}{%
\begin{tabular}{lcc}
\toprule
\textbf{Metric} & \textbf{Base Model} & \textbf{\method} \\
\midrule
Parameters (B)
& $\sim$7 
& +0.05 ($+0.7\%$) \\
TFLOPs 
& 9.24 
& +0.06 ($+0.6\%$) \\
\bottomrule
\end{tabular}%
}\vspace{-0.8em}
\caption{\textbf{Compute cost and parameter count}: \method introduces negligible new parameters and computational cost, while maintaining competitive performance.}\label{tab:compute_comparison}
\vspace{-1.5em}
\end{table}

\section{\dataset and \benchmark Datasets}

\begin{figure*}
    \centering
    \includegraphics[width=0.78\linewidth]{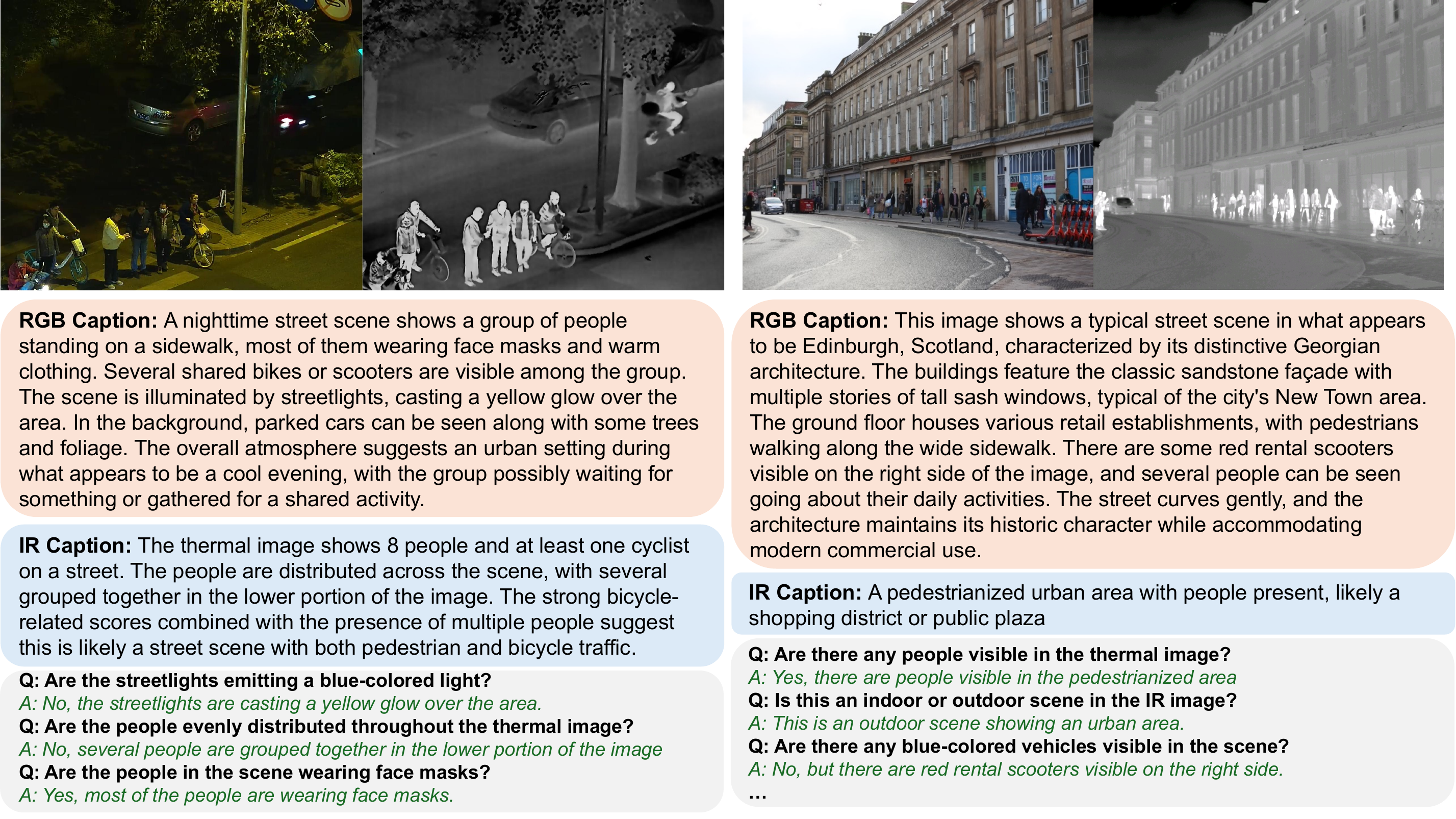}\vspace{-0.8em}
    \caption{\textbf{RGB, IR Captions and QA Pairs.} We show examples of paired RGB–IR images along with their modality-specific captions and the corresponding question–answer pairs generated from those captions. The RGB caption includes richer scene details such as lighting, clothing, and context, while the IR caption reflects high level information like presence of people and overall scene type.}\vspace{-1.5em}
    \label{fig:dataset_examples}
\end{figure*}

To train and evaluate \method, we further introduce \dataset and \benchmark, two comprehensive resources for advancing multimodal understanding across infrared and RGB imagery. \dataset is an instruction-tuning dataset of IR–RGB image pairs with fine-grained question–answer annotations, while \benchmark is a curated benchmark set for assessing modality-specific reasoning and robustness under controlled degradations. Importantly, \dataset is created automatically using an agentic annotation framework.

\subsection{Agentic Framework for IR-RGB Annotation}
Building \method requires a way to generate rich, modality-grounded annotations associated with IR-RGB image pairs. These textual annotations must accurately reflect the unique characteristics of IR-RGB perception. A central obstacle is that IR imagery lacks large-scale captioning and QA resources, unlike the RGB domain where such datasets are abundant. Existing efforts~\cite{jiang_infrared-llava_2024} often sidestep this limitation by synthesizing IR data from RGB images and heuristically adapting RGB captions. However, this strategy relies on imagined thermal appearances rather than genuine measurements, leading to annotations that could misrepresent IR-specific cues such as temperature gradients, emissivity patterns, and heat-based contrast. These systematic biases highlight the need for a more principled approach to annotation grounded in real IR data.

\smallskip
\noindent \textbf{Agentic Framework for Captioning.} To address this gap and inspired by~\cite{ashutosh_llms_2025}, we introduce a modality-aware \textit{agentic annotation framework} in which LLMs act as agents that iteratively generate and refine annotations under the supervision of a pretrained contrastive model. Crucially, unlike prior work that re-purposes RGB captions, our approach operates directly on real IR imagery, ensuring annotations reflect modality-specific content. The pipeline (illustrated in the Supplement), proceeds in three stages:
\begin{enumerate}
    \item \emph{Candidate Generation:} An LLM (Claude Sonnet 3.5 v2~\cite{anthropic_claude}) generates a diverse set of candidate captions based on minimal input cues or object-detection labels when available.  
    \item \emph{Contrastive Refinement:} A modality-specific contrastive model (IR LanguageBind~\cite{zhu_languagebind_2023}) evaluates the candidate text–IR image alignment and assigns similarity scores. These scores guide the LLM through multiple rounds (9 in our case) of iterative refinement in a closed-loop process, progressively improving accuracy and relevance.  
    \item \emph{Final Selection:} Finally, the refined candidate captions, along with their contrastive scores, are synthesized by a stronger LLM (Claude Opus) into a final caption that best captures the semantics of the IR input image.  
\end{enumerate}

To strike a favorable balance between performance and compute, we employ the more efficient Claude Sonnet during the multi-iteration refinement loop and reserve Claude Opus for a single final reasoning pass. Empirically, this configuration provides a strong accuracy–efficiency trade-off. Using this framework, we automatically generate captions for $\sim$25K IR images in LLVIP~\cite{jia_llvip_2021} and HDRT~\cite{peng_hdrt_2025}.
\begin{figure*}[th!]
    \centering
    \includegraphics[width=0.8\linewidth]{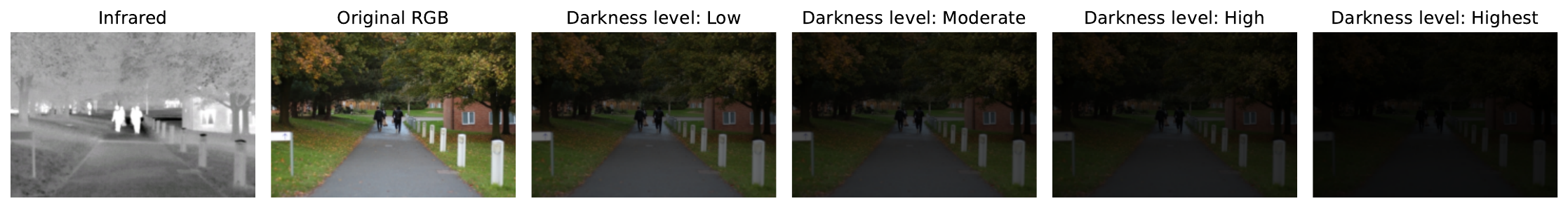}\vspace{-2mm}
    \includegraphics[width=0.8\linewidth]{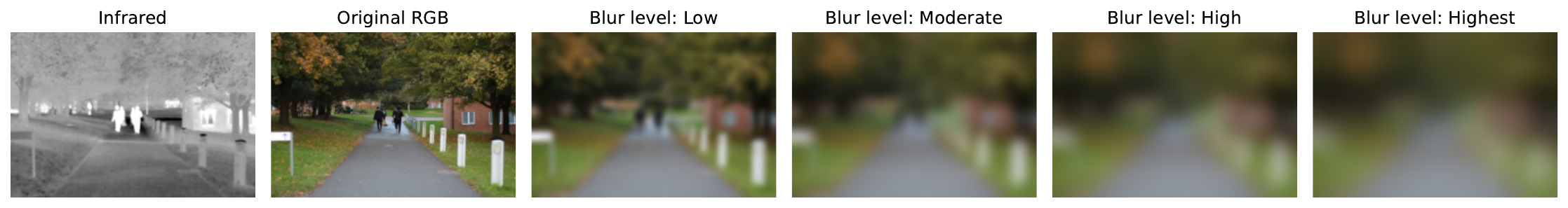}\vspace{-2mm}
    \includegraphics[width=0.8\linewidth]{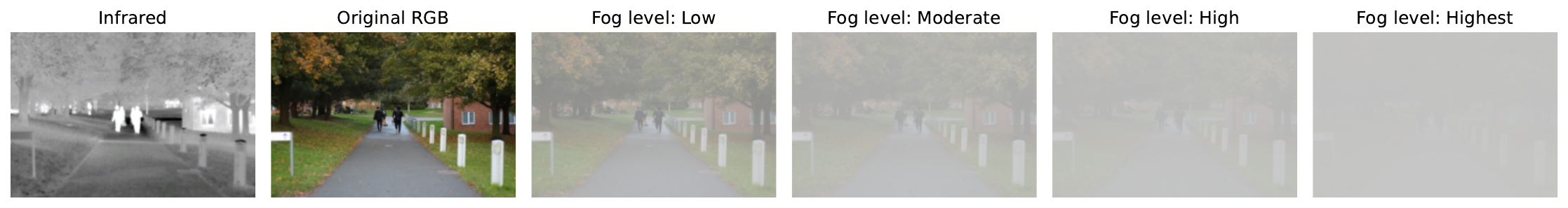}\vspace{-1em}
    \caption{\textbf{RGB degradation conditions in \benchmark.} Examples of the three corruption types: darkness, blur, and fog, applied at four severity levels each, with IR left unaltered. These controlled degradations enable systematic evaluation of the robustness of VLMs.}
\label{fig:effects}\vspace{-1.5em}
\end{figure*}

\smallskip
\noindent \textbf{Assessing the Quality of Generated Captions}. We further evaluate the quality of captions from our agentic framework. Directly evaluating IR captions is challenging due to the lack of ground-truth benchmarks for infrared imagery. To address this, we apply an \textit{LLM-as-a-judge} evaluation on a random subset of $\sim$1700 paired IR–RGB images. 

This is achieved by the following steps. \textit{First}, leveraging the strong RGB captioning of modern MLLMs, high-quality RGB reference captions are generated (using Claude Sonnet 3.5v2). \textit{Second}, a judging LLM (a different instance of Sonnet 3.5v2) then scores the corresponding IR captions against these references while being instructed to account for modality-specific visibility differences. Each IR caption is rated along two dimensions: \emph{Accuracy} (faithfulness to IR-visible content) and \emph{Detail} (descriptive completeness).

\begin{table}[t]
\centering
\resizebox{0.95\columnwidth}{!}{%
\begin{tabular}{lcccc}
\toprule
& \multicolumn{4}{c}{\textbf{Score}} \\
\cmidrule(lr){2-5}
\textbf{Metric} & \textbf{Very Good} & \textbf{Good} & \textbf{Fair} & \textbf{Poor} \\
\midrule
Accuracy & 873 (51.1\%) & 609 (35.6\%) & 164 (9.6\%) & 64 (3.7\%) \\
Detail   & 2 (0.1\%)    & 807 (47.2\%) & 807 (47.2\%) & 94 (5.5\%)  \\
\bottomrule
\end{tabular}}\vspace{-0.5em}
\caption{\textbf{Quality of IR captions} is evaluated using an LLM-as-a-judge protocol, comparing IR captions against reference RGB captions from paired images. Most IR captions score \textit{Good} or better in accuracy, while descriptive detail is lower, consistent with the reduced visual cues in infrared imagery.
}\label{tab:cap_eval}\vspace{-1.5em}
\end{table}

As summarized in Table \ref{tab:cap_eval}, IR captions demonstrate strong overall quality. Most (86.7\%) achieve ``Good'' or ``Very Good'' ratings for accuracy, showing that they reliably capture IR scene content. About half reach similar ratings for detail, as IR imagery inherently provides less fine-grained information than RGB. These results confirm that our framework produces informative and faithful IR descriptions directly from the IR images.

\smallskip
\noindent \textbf{From Captions to QA Pairs.} Finally, we convert each caption into 2-4 QA pairs using an LLM (Claude Sonnet 3.5 v2). Every QA pair is tagged with the modality (IR or RGB) that provides the key visual evidence, enabling modality-aware supervision. RGB questions target fine-grained appearance cues (\eg, ``\emph{What color is the car?}''), while IR questions focus on properties reliably visible in infrared, such as object count or coarse scene layout (\eg, ``\emph{How many people are visible?}''). 
Sample IR-RGB images, their captions and converted QA pairs are shown in Fig.~\ref{fig:dataset_examples}.

\subsection{\dataset for Instruction Tuning}
We applied our annotation pipeline to $\sim$9.5K aligned IR–RGB image pairs from the HDRT dataset~\cite{peng_hdrt_2025} (reserving 500 pairs for testing) and an additional $\sim$15K pairs from the LLVIP~\cite{jia_llvip_2021} dataset. These two sources are complementary: HDRT captures diverse environmental settings, while LLVIP emphasizes low-light urban scenes.

Our \emph{\dataset} dataset contains $\sim$25K aligned IR–RGB pairs with $\sim$204K QA annotations, averaging $8.1$ QA pairs per image. Following~\cite{liu_improved_2024}, QA pairs are formulated in an open-ended, instruction-tuning format (\eg, ``What color is the car?'' $\rightarrow$ ``The car is blue.''). Importantly, questions are evenly divided between RGB- and IR- dependent cues, ensuring that learned models are exposed to both modality-specific reasoning and cross-modal alignment. 

\textit{\dataset is intended for instruction tuning of MLLMs}, offering large-scale IR-RGB data for modality-aware reasoning and robust multimodal integration.

\subsection{\benchmark for Evaluation}
A key goal of IR–RGB fusion is maintaining accuracy when the RGB modality is degraded. To evaluate this behavior under controlled conditions, we introduce \benchmark, a dataset of 500 aligned IR–RGB image pairs paired with 500 QA items. Each question is constructed to require complementary information from both modalities, semantic structure from IR and fine-grained appearance cues from RGB. This design allows us to assess if IR images can compensate for degraded RGB images,  while still leveraging any remaining texture or color information.

For objective scoring, open-ended questions are converted into binary yes/no statements, following~\cite{ashutosh_llms_2025,nguyen_owlviz_2025}. Images are sampled from HDRT for high-quality RGB references and degradation is applied.
To simulate visual degradation while preserving IR signals, we apply corruptions only to RGB images, since long-wave infrared sensors measure emitted thermal radiation and are largely unaffected by many visible-light degradations. Darkness does not alter thermal emission; optical blur primarily impacts visible-spectrum optics; and fog or haze scatters short wavelengths far more than long-wave IR. However, we acknowledge that this setup assumes idealized IR robustness under these conditions. We implement three corruption types (\emph{darkness}, \emph{blur}, and \emph{fog}), each at four severity levels, plus a clean condition, yielding 13 evaluation categories (Fig.~\ref{fig:effects}). Details are provided in the Supplement.

\textit{\benchmark is intended for evaluation}, revealing how well models integrate complementary IR and RGB cues under diverse degradation types and severity levels. %

\section{Experiments and Results}

We present a comprehensive evaluation of \method. We outline the baselines and evaluation metrics, followed by systematic ablations that examine the impact of IR-RGB fusion and the effects of model design choices. These analyses establish the final version of \method used for comparison with state-of-the-art MLLMs.

\begin{figure*}[th!]
    \centering
    \includegraphics[width=0.75\linewidth]{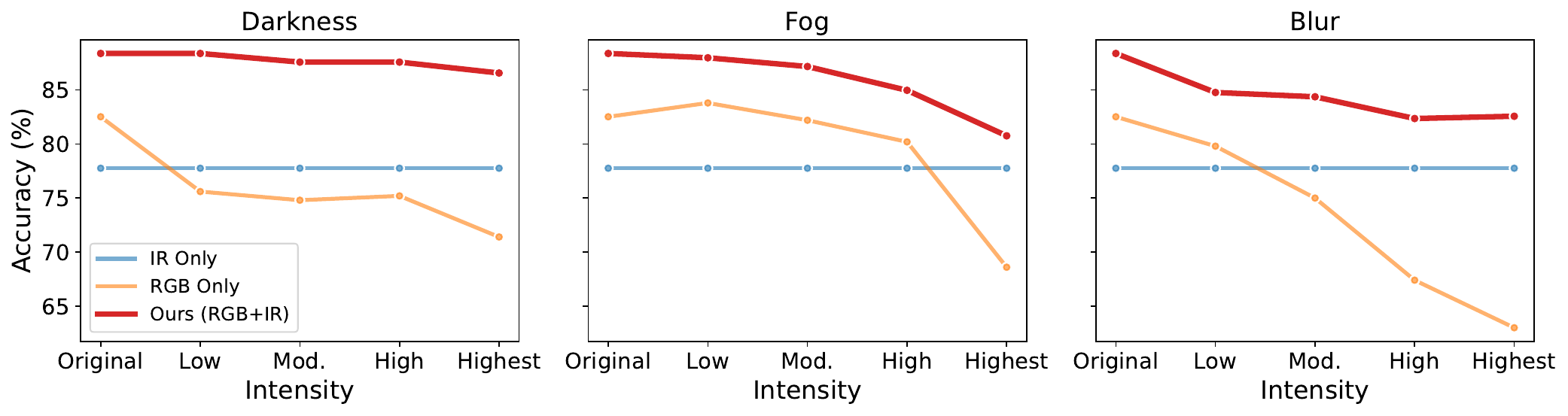}\vspace{-1.1em}
    \caption{\textbf{Performance by Modalities (IR, RGB, RGB+IR)}. IR-only performance stays flat across degradations, RGB-only performance drops sharply as degradation severity increases, and RGB+IR provides the most robust results. Full results are in the Supplement.}\vspace{-1.em}
    \label{fig:modality_ablation}
\end{figure*}

\begin{figure*}
    \centering
    \includegraphics[width=0.75\linewidth]{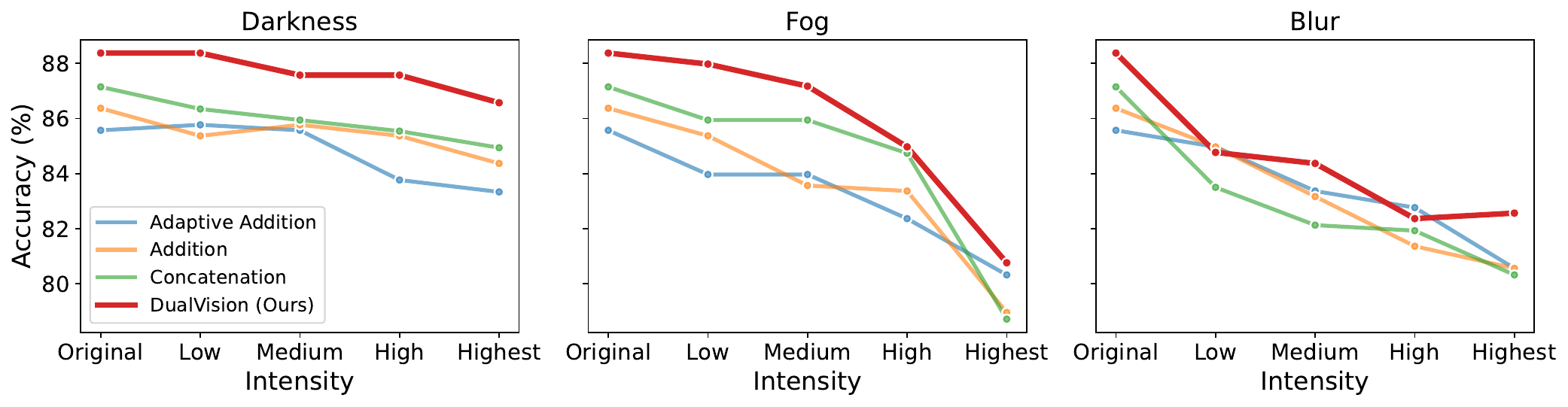}\vspace{-1.1em}
    \caption{\textbf{Comparison of Fusion Methods.} We compare several fusion strategies: addition, adaptive addition, concatenation, and our \method. Note that the concatenation baseline is equivalent to finetuned LLaVA-1.5-7B. Full results can be found in the Supplement.}\vspace{-1.5em}
    \label{fig:results_plot}
\end{figure*}

\smallskip
\noindent \textbf{Baselines.} We benchmark against both open- and closed-source MLLMs. Open-source baselines include LLaVA 1.5–7B~\cite{liu_improved_2024}, Qwen2-VL 7B~\cite{wang_qwen2-vl_2024}, LLaVA-Next Interleave 7B~\cite{li_llava-next-interleave_2024}, and LLaMA-4 Scout~\cite{llama_4}. For completeness, we also evaluate some closed-source commercial systems such as Anthropic Claude Sonnet 3.5v2 and Claude Opus $4$~\cite{anthropic_claude}. All baselines are evaluated with RGB and IR images at a resolution of 336$\times$336 to ensure a fair comparison.

To study the role of IR-RGB fusion, we evaluate LLaVA~1.5-7B variants using IR-only, RGB-only, and their combination under identical conditions. To assess the effectiveness of our fusion design, we implement several popular fusion strategies, each trained on $\dataset$ with identical hyperparameters and the same degradation-aware protocol. These include token-wise addition of embeddings, adaptive weighted addition (where learnable token weights are applied before summation), and the commonly used concatenation (interleaving tokens), followed by linear projection. All fusion variants share the LLaVA~1.5-7B backbone and the frozen CLIP ViT-$L/14$ encoder $E$, ensuring that performance differences arise solely from the fusion mechanism.

\smallskip
\noindent \textbf{Training Details.} Models are trained for $2$ epochs on the \dataset dataset, which contains both RGB- and IR-oriented QA pairs. Training is performed on $8 \times$ Nvidia A100 GPUs (80 GB RAM) with FlashAttention for memory efficiency, and DeepSpeed 
for further training optimization. We use a per-device batch size of 16 and learning rates of $10^{-4}$ for the fusion module and $10^{-5}$ for $P$, along with linear warm-up followed by cosine annealing. Under this setup, training completes in roughly one hour on \dataset.

\smallskip
\noindent \textbf{Metrics and Evaluation Protocol.} Evaluation is conducted on \benchmark. %
Since answers in \benchmark are binary (\ie, yes/no), exact-match accuracy serves as the primary metric.
We specifically analyze the setting where both modalities are provided and the question requires information from each. In this context, IR contributes stable semantic structure, while RGB supplies fine-grained texture and color cues. This design allows us to probe whether IR can compensate when RGB becomes degraded, while still leveraging whatever residual color or texture information remains in the corrupted RGB input. Following the \benchmark protocol, degradations applied to RGB images include blur, darkness, and fog, each at four severity levels, in addition to clean images. Results are stratified by corruption type and severity to enable fine-grained robustness assessment.

\begin{table*}[th!]
\centering
\resizebox{\textwidth}{!}{%
\begin{tabular}{l c c ccccccccccccc}
\toprule
\textbf{Method} & \textbf{\#Params} & \textbf{\#Tokens} & \textbf{Original} 
& \multicolumn{4}{c}{\textbf{Blur}} 
& \multicolumn{4}{c}{\textbf{Darkness}} 
& \multicolumn{4}{c}{\textbf{Fog}} \\
\cmidrule(lr){4-4} \cmidrule(lr){5-8} \cmidrule(lr){9-12} \cmidrule(lr){13-16}
 & & & & Low & Mod. & High & Highest & Low & Mod. & High & Highest & Low & Mod. & High & Highest \\
\midrule
\multicolumn{16}{l}{\textbf{w/o Finetuning}} \\
\midrule

LLaVA 1.5–Text Only~\cite{liu_improved_2024} 
  & 7B & -- & 48.8 & -- & -- & -- & -- & -- & -- & -- & -- & -- & -- & -- \\

LLaVA 1.5~\cite{liu_improved_2024}           
  & 7B & 2N & 81.2 & 79.0 & 76.0 & 73.2 & 72.6 & 74.0 & 72.8 & 73.2 & 71.2 & 81.0 & 79.2 & 78.4 & 71.4 \\

Qwen2-VL~\cite{wang_qwen2-vl_2024}               
  & 7B & 2N & 89.8 & 77.8 & 73.6 & 70.8 & 69.4 & 89.6 & 85.6 & 82.6 & 78.4 & 85.0 & 79.4 & 75.2 & 65.6 \\

Qwen2.5-VL~\cite{bai2025qwen25}               
  & 7B & 2N & \textbf{90.2} 
& 75.0 & 65.0 & 61.0 & 60.6 
& \textbf{89.4} & 83.8 & 80.6 & 72.8 
& 85.8 & 78.2 & 71.4 & 61.2 \\

LLaVA-Next Interleave Qwen~\cite{li_llava-next-interleave_2024}
  & 7B & 2N & 88.6 & 81.4 & 78.4 & 75.2 & 73.6 & 86.4 & 86.0 & 85.6 & 81.4 & 85.0 & 83.8 & 79.8 & 73.4 \\

LLaMA-4 Scout~\cite{llama_4}
  & 17B & 2N & 85.8 & 71.6 & 62.6 & 59.2 & 57.6 & 79.4 & 71.6 & 66.4 & 56.6 & 73.4 & 63.6 & 58.0 & 54.0 \\

\cdashline{1-16}

Claude Opus 4~\cite{anthropic_claude}
  & -- & 2N & 86.6 & 67.0 & 60.8 & 60.0 & 58.8 & 83.6 & 72.4 & 64.6 & 57.0 & 74.2 & 63.8 & 59.2 & 56.2 \\

Claude Sonnet-3.5 v2~\cite{anthropic_claude}
  & -- & 2N & 87.4 & 77.8 & 74.8 & 70.8 & 72.0 & 85.4 & 78.8 & 75.6 & 68.0 & 80.0 & 70.2 & 68.4 & 64.4 \\

\midrule
\multicolumn{16}{l}{\textbf{Finetuned}} \\
\midrule

LLaVA 1.5~\cite{liu_improved_2024}
  & 7B & 2N & 87.15 & 83.50 & 82.13 & 81.93 & 80.32 & 86.35 & 85.94 & 85.54 & 84.94 & 85.94 & 85.94 & 84.74 & 78.71 \\

\method (ours)
  & 7B & \textbf{N} & 88.38 & \textbf{84.77} & \textbf{84.37} & \textbf{82.36} & \textbf{82.57} 
    & 88.38 & \textbf{87.58} & \textbf{87.58} & \textbf{86.57} 
    & \textbf{87.98} & \textbf{87.17} & \textbf{84.97} & \textbf{80.76} \\
\bottomrule
\end{tabular}}\vspace{-0.5em}
\caption{\textbf{Main results on \benchmark}. We compare our method against strong open-source and commercial MLLM baselines under clean and corrupted conditions. Our approach achieves the highest overall accuracy and shows improved robustness against degradations. }\label{tab:baselines}\vspace{-0.5em}
\end{table*}

\begin{figure*}[th!]
    \centering
    \includegraphics[width=0.9\linewidth]{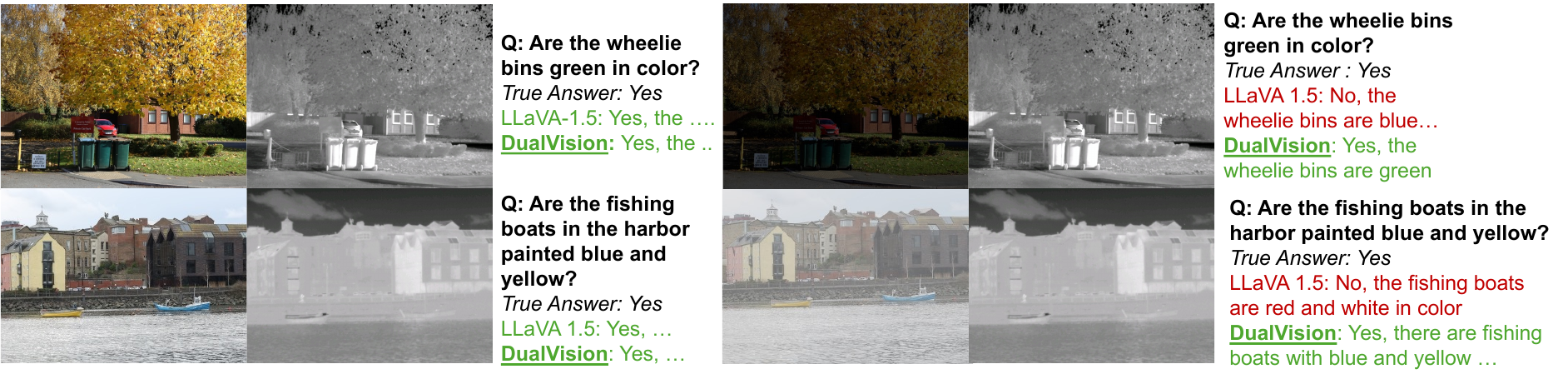}\vspace{-1.2em}
    \caption{\textbf{Results on \benchmark.} Both methods, finetuned on \dataset, answer correctly with clean RGB–IR inputs (left). When the RGB is degraded (right), we can see \method remains robust, while finetuned LLaVA-1.5 shows reduced reliability.}\vspace{-1.5em}
    \label{fig:sample_results}
\end{figure*}

\subsection{Design and Ablation Study}

\noindent \textbf{Effects of Modalities}. 
We first analyze the contribution of each modality to overall robustness. The RGB-only model is evaluated using the pretrained LLaVA~1.5 7B weights without additional finetuning, while the IR-only variant is finetuned on \dataset using only IR images for fairness. As in Fig.~\ref{fig:modality_ablation}, single-modality models perform substantially worse under both clean and degraded settings. The IR-only model attains the lowest accuracy on \benchmark, and its performance remains flat across degradations since no RGB input is available. The RGB-only model performs well on clean inputs but deteriorates rapidly as corruption severity increases. In contrast, integrating both modalities yields consistent improvements across all degradation types. The widening gap between the RGB-only model and our approach under degradation highlights the importance of multimodal fusion for robust perception.

\smallskip
\noindent \textbf{Fusion Design}. We compare \method with several established IR-RGB fusion methods to assess its effectiveness in multimodal integration. Specifically, we implement three canonical fusion strategies: token-wise addition, adaptive weighted addition, and concatenation, each finetuned on \dataset with identical hyperparameters and the same degradation-aware training protocol. All models share the LLaVA~1.5--7B backbone and frozen CLIP encoder $E$, ensuring that performance differences arise solely from the fusion mechanism. As illustrated in Fig.~\ref{fig:results_plot}, \method delivers the best performance, winning 
$11/13$ evaluation settings over both clean and degraded RGB inputs. Detailed results can be found in the Supplement.

\smallskip
\noindent \textbf{Additional Ablation Studies}. We also conduct detailed analyses of both our attention design and degradation-aware training. More details are provided in the Supplement.

\subsection{Comparison with Baselines}

Table~\ref{tab:baselines} summarizes results for the baselines, including open-source VLMs (LLaVA variants, Qwen2-VL) and large closed-source systems. As a sanity check, we also evaluate LLaVA~1.5 7B without providing any visual input, and it performs at chance level (48.8\%), confirming dataset balance. When provided with RGB and IR inputs, all VLMs achieve strong performance on clean data, with Qwen2-VL, LLaVA-Next, and closed-source models consistently reaching the mid-to-high $80\%$ range. This demonstrates that modern MLLMs can effectively interpret visual information under ideal conditions.
However, performance drops sharply under degradations such as blur, darkness, or fog, with most baselines losing $15–20$ percentage points in accuracy. Finetuning LLaVA~1.5 7B on \dataset using our degradation-aware protocol substantially improves robustness, confirming the benefit of training under degraded visual conditions. Nonetheless, \method consistently outperforms this finetuned baseline and all other VLMs across most settings. These results highlight that while degradation-aware learning enhances general robustness, the fusion and attention mechanisms in \method remain crucial for achieving state-of-the-art multimodal resilience. Qualitative comparisons in Figure~\ref{fig:sample_results} further illustrate \method's superior ability to preserve semantic fidelity under challenging visual conditions.

\section{Conclusion}
This paper presents one of the first efforts to develop MLLMs that integrate IR and RGB images for robust visual perception and reasoning under various visual degradations. We introduced \method, a lightweight IR-RGB fusion module for MLLMs that performs multi-scale localized cross-attention, enabling effective interaction between modalities.
Together with our \dataset for instruction tuning and \benchmark for evaluation, \method establishes an effective solution for IR-RGB reasoning. Our experiments demonstrated that the performance of current MLLMs drops significantly under adverse degradations (\eg, blur, low light, and fog), whereas \method delivers consistent improvements across all settings. Future work could investigate end-to-end finetuning, misalignment-tolerant fusion, and broader multimodal corruption benchmarks, along with real-world evaluations.

\noindent\textbf{Acknowledgment}. This material is partially based on work supported by the Army Research Office funded Assured Autonomy Innovation Institute (A2I2) under Contract number W911NF-2020-221, and the National Science Foundation under Grant Number CNS-2333491. Any opinions, findings, and conclusions or recommendations expressed in this material are those of the authors and do not necessarily reflect the views of the sponsors.

{
    \small
    \bibliographystyle{ieeenat_fullname}
    \bibliography{DualVision}
}

\clearpage
\setcounter{figure}{0}
\setcounter{table}{0}
\setcounter{section}{0}
\setcounter{equation}{0}
\renewcommand{\thefigure}{\Alph{figure}}
\renewcommand{\thesection}{\Alph{section}}
\renewcommand{\thetable}{\Alph{table}}
\renewcommand{\theequation}{\Alph{equation}}
\setcounter{page}{1}

\maketitlesupplementary

This supplementary document provides expanded experimental results and analyses that complement the main paper. We present additional ablations that examine our design (Section \ref{sec:supp_ablation}), provide further quantitative and qualitative results omitted from the main paper due to space limits (Section \ref{sec:supp_results}), introduce details about the construction of our datasets (Sec.\ \ref{sec:supp_datasets}), and describe implementation details of \method (Section \ref{sec:supp_impl}).

For sections, figures and equations, we use numbers (\eg, Sec.\ 1) to refer to the main paper and capital letters (\eg, Sec.\ A) to refer to this supplement.

\section{Additional Ablations}\label{sec:supp_ablation}
We present ablation studies examining the attention design of \method as well as the contribution of degradation-aware training. These experiments highlight the effectiveness of our design choices and demonstrate their importance for achieving robust multimodal fusion.

\smallskip
\noindent \textbf{Attention Design.}  We dissect the contributions of the central design choice \ie, the attention mechanism design. Specifically, we compare global cross-attention, fixed-radius local attention (\ie, $r=1$), and our multi-scale local attention. As shown in Table~\ref{tab:ablation_attention}, the multi-scale variant achieves the best overall performance, outperforming alternatives in $8$ of $12$ degraded settings. Fixed-radius attention occasionally matches or slightly exceeds our approach on clean data ($88.58\%$ vs.\ $88.38\%$), suggesting that simpler fusion may suffice under ideal conditions. However, as degradations intensify, multi-scale attention consistently demonstrates stronger robustness, validating the hypothesis that localized interactions enable more effective cross-modal integration.

\smallskip
\noindent \textbf{Degradation-Aware Training.}  
We assess the role of training with stochastic degradations. Table~\ref{tab:ablation_training} shows that degradation-aware training universally improves performance. Under severe blur, \method improves by +6.47\% (82.57\% vs.\ 76.10\%); under the highest darkness level, by +4.81\%; and under severe fog, also by +4.81\%. These results demonstrate that exposure to corrupted inputs during training equips the model with more resilient fusion behaviors, enabling better generalization under adverse visual conditions.

\begin{table*}[hbtp!]
\centering
\resizebox{\textwidth}{!}{%
\begin{tabular}{l c c cccc cccc cccc}
\toprule
\textbf{Method} & \textbf{Wins} & \textbf{Original}
& \multicolumn{4}{c}{\textbf{Blur}}
& \multicolumn{4}{c}{\textbf{Darkness}}
& \multicolumn{4}{c}{\textbf{Fog}} \\
\cmidrule(lr){4-7}\cmidrule(lr){8-11}\cmidrule(lr){12-15}
& & & Low & Moderate & High & Highest & Low & Moderate & High & Highest & Low & Moderate & High & Highest \\
\midrule
Global xAttention            & 1 & 88.18 & 83.94 & 81.96 & 81.76 & 80.96 & 87.37 & 87.58 & \textbf{87.58} & 85.57 & 86.17 & 86.37 & 84.37 & 78.92 \\
Local xAttention ($r{=}1$) & 5 & \textbf{88.58} & \textbf{84.97} & 83.37 & \textbf{83.17} & 81.96 & \textbf{88.78} & \textbf{87.78} & 87.17 & 85.97 & 87.17 & 86.17 & 84.17 & 80.56 \\
\method (ours)            & \textbf{8} & 88.38 & 84.77 & \textbf{84.37} & 82.36 & \textbf{82.57} & 88.38 & 87.58 & \textbf{87.58} & \textbf{86.57} & \textbf{87.98} & \textbf{87.17} & \textbf{84.97} & \textbf{80.76} \\
\bottomrule
\end{tabular}}\vspace{-0.5em}
\caption{\textbf{Ablation on Attention Design}. Accuracy (\%) is reported across corruptions on \benchmark. All models use 3 blocks. \emph{Global xAttn} uses full cross-attention; \emph{Local xAttn ($r=1$)} uses fixed local neighborhoods; \method applies multi-scale local xAttn $(r\in \{1,2,3\})$.
}\label{tab:ablation_attention}
\end{table*}

\begin{table*}[hbtp!]
\centering
\resizebox{\textwidth}{!}{%
\begin{tabular}{l c c cccc cccc cccc}
\toprule
\textbf{Method} & \textbf{Wins} & \textbf{Original}
& \multicolumn{4}{c}{\textbf{Blur}}
& \multicolumn{4}{c}{\textbf{Darkness}}
& \multicolumn{4}{c}{\textbf{Fog}} \\
\cmidrule(lr){4-7}\cmidrule(lr){8-11}\cmidrule(lr){12-15}
& & & Low & Moderate & High & Highest & Low & Moderate & High & Highest & Low & Moderate & High & Highest \\
\midrule
DualVision (w/o Deg.-Aware Training) & 0  & 88.18 & 84.17 & 79.56 & 76.91 & 76.10 & 86.97 & 85.57 & 84.17 & 81.76 & 86.57 & 83.97 & 81.76 & 75.95 \\
DualVision (w/  Deg.-Aware Training) & \textbf{13} & \textbf{88.38} & \textbf{84.77} & \textbf{84.37} & \textbf{82.36} & \textbf{82.57} & \textbf{88.38} & \textbf{87.58} & \textbf{87.58} & \textbf{86.57} & \textbf{87.98} & \textbf{87.17} & \textbf{84.97} & \textbf{80.76} \\
\bottomrule
\end{tabular}}\vspace{-0.5em}
\caption{\textbf{Ablation on Degradation-Aware Training.} Accuracy (\%) is reported across corruption types and severities on DualVision-500.}
\label{tab:ablation_training}
\end{table*}

\begin{table*}[hbtp!]
\centering
\resizebox{\textwidth}{!}{%
\begin{tabular}{l c ccccccccccccc}
\toprule
\textbf{Method} & \textbf{Original} 
& \multicolumn{4}{c}{\textbf{Blur}} 
& \multicolumn{4}{c}{\textbf{Darkness}} 
& \multicolumn{4}{c}{\textbf{Fog}} \\
\cmidrule(lr){2-2} \cmidrule(lr){3-6} \cmidrule(lr){7-10} \cmidrule(lr){11-14}
 &  & Low & Mod. & High & Highest & Low & Mod. & High & Highest & Low & Mod. & High & Highest \\
\midrule
LLaVA 1.5 7B-IR Only (Finetuned) & 77.76 & - & - & - & - & - & - & - & - & - & - & - & - \\
LLaVA 1.5 7B-RGB Only (OOB)      & 82.52 & 79.80 & 75.00 & 67.40 & 63.00 & 75.60 & 74.80 & 75.20 & 71.40 & 83.80 & 82.20 & 80.20 & 68.60 \\
\method (ours)                   & \textbf{88.38} & \textbf{84.77} & \textbf{84.37} & \textbf{82.36} & \textbf{82.57} & \textbf{88.38} & \textbf{87.58} & \textbf{87.58} & \textbf{86.57} & \textbf{87.98 }& \textbf{87.17} & \textbf{84.97} & \textbf{80.76} \\
\bottomrule
\end{tabular}}\vspace{-0.5em}
\caption{\textbf{Ablation on Different Modalities}. Accuracy (\%) is reported across corruption types and severities on DualVision-500. All models use the LLaVA~1.5 7B~\cite{liu_improved_2024} backbone. }
\label{tab:modality_ablation}
\end{table*}

\begin{table*}[hbtp!]
\centering
\resizebox{\textwidth}{!}{%
\begin{tabular}{l c ccccccccccccc}
\toprule
\textbf{Method} & \textbf{Wins} & \textbf{Original} 
& \multicolumn{4}{c}{\textbf{Blur}} 
& \multicolumn{4}{c}{\textbf{Darkness}} 
& \multicolumn{4}{c}{\textbf{Fog}} \\
\cmidrule(lr){3-3} \cmidrule(lr){4-7} \cmidrule(lr){8-11} \cmidrule(lr){12-15}
 & & & Low & Mod. & High & Highest & Low & Mod. & High & Highest & Low & Mod. & High & Highest \\
\midrule
Addition                           & 1  & 86.37 & 84.97 & 83.17 & 81.36 & 80.56 & 85.37 & 85.77 & 85.37 & 84.37 & 85.37 & 83.57 & 83.37 & 78.96 \\
Adaptive Addition                  & 2  & 85.57 & \textbf{84.97} & 83.37 & \textbf{82.77} & 80.56 & 85.77 & 85.57 & 83.77 & 83.33 & 83.97 & 83.97 & 82.36 & 80.32 \\
Concatenation$\dagger$      & 0  & 87.15 & 83.50 & 82.13 & 81.93 & 80.32 & 86.35 & 85.94 & 85.54 & 84.94 & 85.94 & 85.94 & 84.74 & 78.71 \\
\method (ours)             & \textbf{11} & \textbf{88.38} & 84.77 & \textbf{84.37} & 82.36 & \textbf{82.57} & \textbf{88.38} & \textbf{87.58} & \textbf{87.58} & \textbf{86.57} & \textbf{87.98} & \textbf{87.17} & \textbf{84.97} & \textbf{80.76} \\
\bottomrule
\end{tabular}}\vspace{-0.5em}
\caption{\textbf{Ablation on Fusion Strategies}. Accuracy (\%) is reported across corruption types and severities on DualVision-500. Note: All methods are finetuned with the same training data, settings as well as degradation aware training protocol. $\dagger$Equivalent to LLaVA1.5 7B~\cite{liu_improved_2024} finetuned on \dataset with interleaved RGB and IR tokens. }
\label{tab:fusion_results}
\end{table*}

\section{Detailed Results}\label{sec:supp_results}

We provide the detailed numerical results corresponding to the modality ablation and fusion experiments shown in Figures 5 and 6 of the main paper. These tables list the exact accuracy values used to generate the plots, covering all degradation types and severity levels.

\smallskip
\noindent \textbf{Effect of Modalities}.
Table \ref{tab:modality_ablation} reports the detailed accuracy of the RGB-only, IR-only, and RGB–IR variants evaluated in Figure 5 of the main paper, covering all blur, darkness, and fog levels. The results detail how each model responds as degradations intensify, showing the characteristic drop in performance for RGB-only reasoning and the limited overall accuracy of IR-only predictions. In aggregate, the combined RGB–IR model maintains the strongest and most stable performance across the full degradation spectrum.

\smallskip
\noindent \textbf{Fusion Design}.
Table \ref{tab:fusion_results} provides the full accuracy breakdown for the fusion mechanisms as illustrated in Figure 6 of the main paper, including simple addition, adaptive weighted addition, concatenation, and our proposed \method. The table shows how each method behaves under clean and progressively degraded RGB inputs, exposing where robustness differences emerge among the baselines. Across all corruption types and severity levels, \method achieves the highest overall performance among the evaluated fusion strategies.

\begin{figure}[t!]
    \centering
    \includegraphics[width=\linewidth]{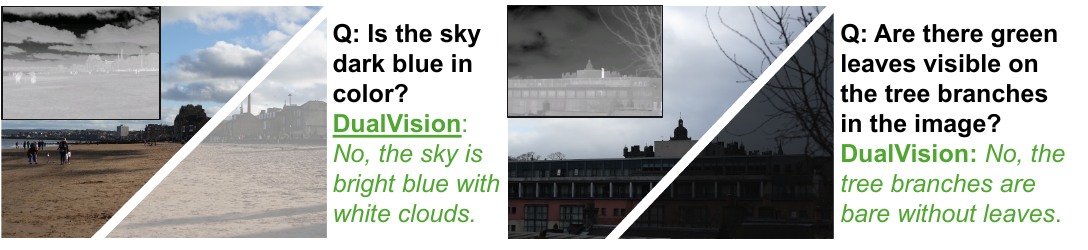}
    \vspace{-2em}
    \caption{Sample results of \method under degradations.}
    \vspace{-1em}
    \label{fig:reasoning}
\end{figure}

\begin{figure*}[ht!]
    \centering
    \vspace{-0.2em}
    \includegraphics[width=0.9\linewidth]{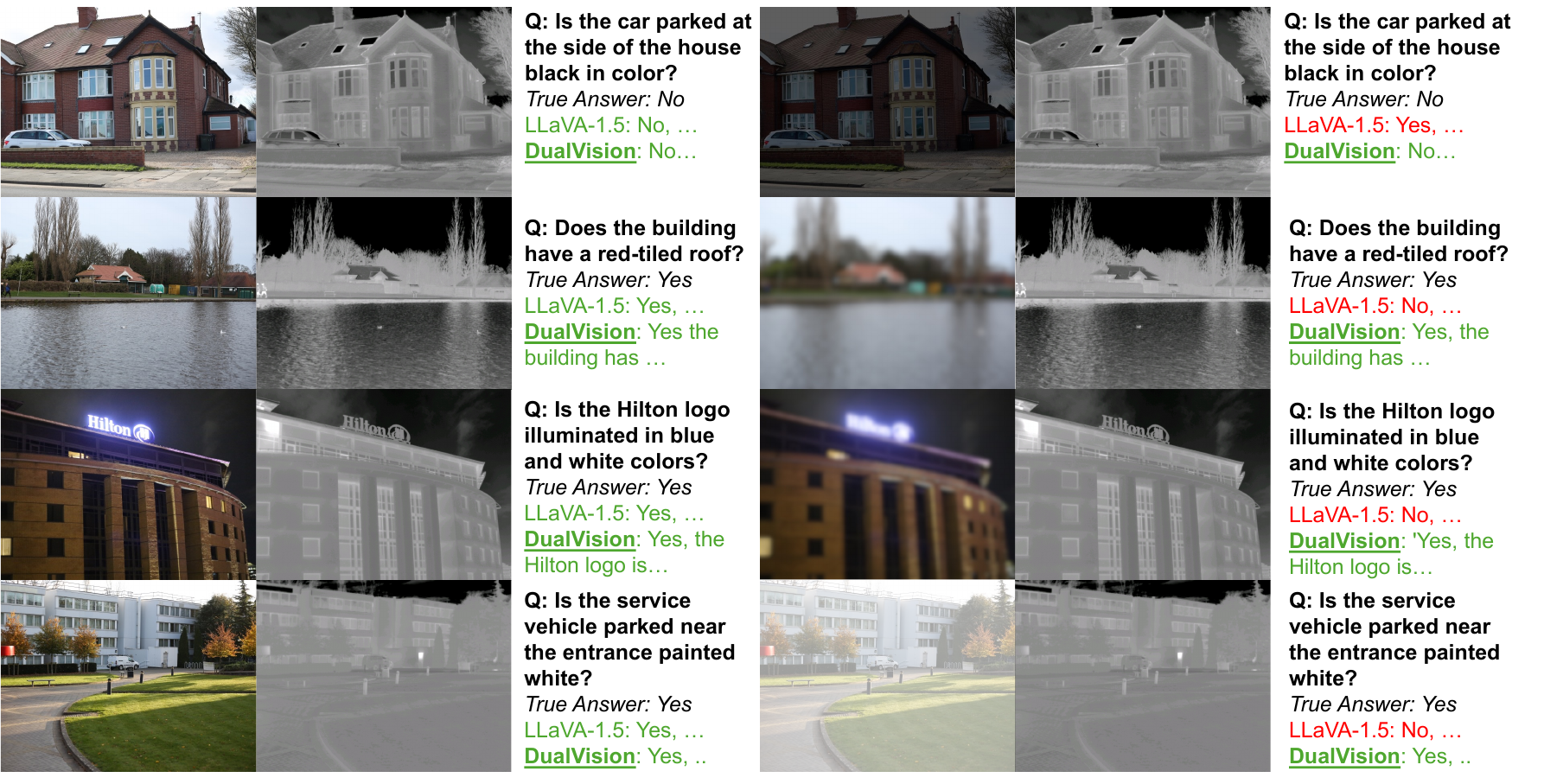}\vspace{-1em}
    \caption{\textbf{Sample results} (from \benchmark) showing how \method maintains accurate predictions across different degradation types.}
    \label{fig:supp_examples}
    \vspace{-1.5em}
\end{figure*}

\smallskip
\noindent \textbf{Additional Qualitative Results}.
Figure~\ref{fig:reasoning} showcases the model’s ability to produce more extended and detailed reasoning in its responses, while Figure~\ref{fig:supp_examples} provides additional examples that highlight the robustness of \method across a range of degradations.

\begin{figure}[ht!]
    \centering
    \includegraphics[width=0.95\linewidth]{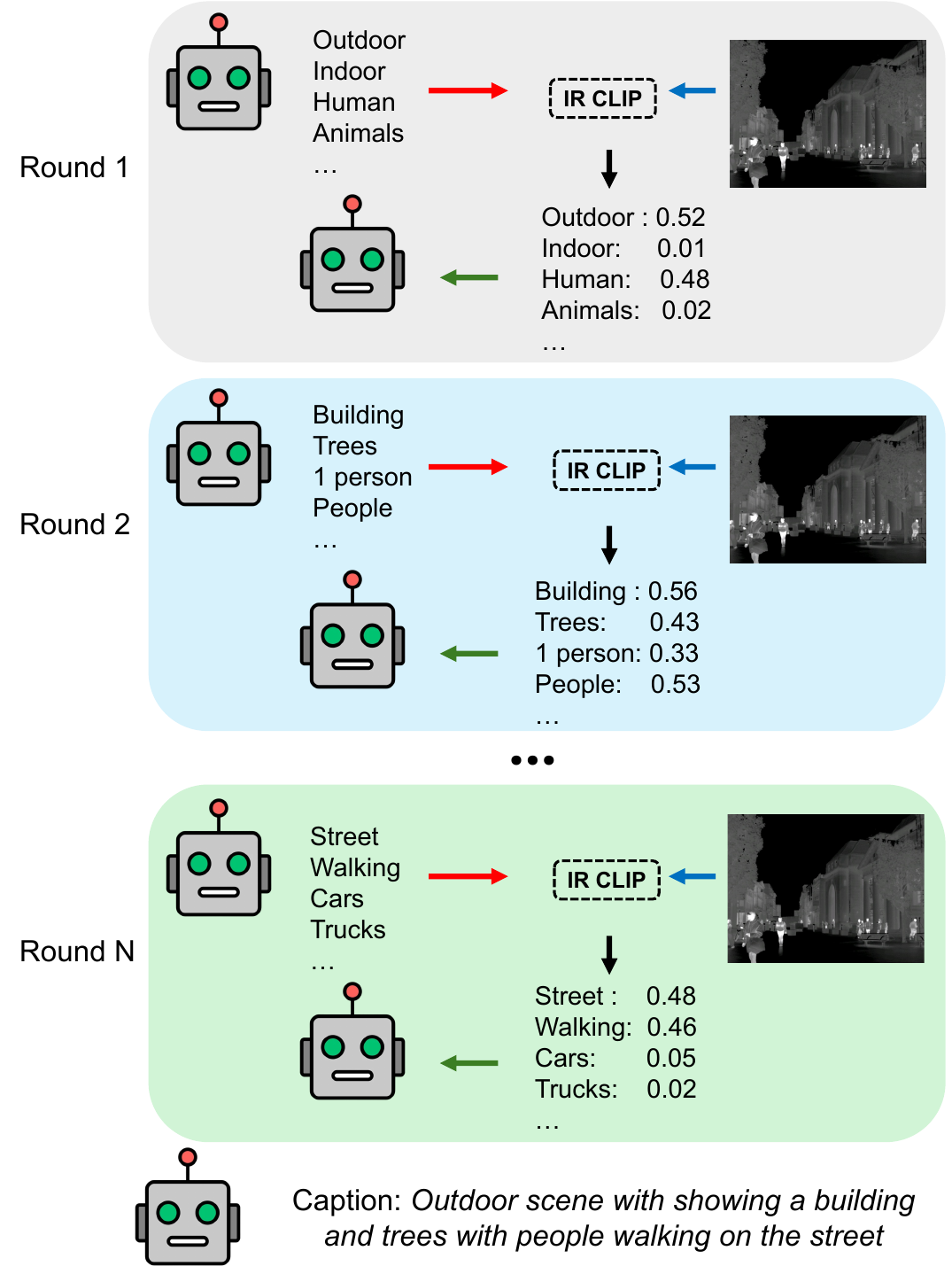}
    \caption{\textbf{Our Agentic Framework} for captioning IR images. An LLM proposes and refines caption candidates while IR-CLIP~\cite{zhu_languagebind_2023} provides similarity-based supervision at each iteration.}
    \label{fig:supp_anno_demo}\vspace{-1.5em}
\end{figure}

\section{Dataset Details}\label{sec:supp_datasets}

We illustrate the agentic annotation framework and provide additional details on the degradation simulation procedures used throughout our experiments.

\subsection{Agentic Framework for Captioning}
Our three-stage annotation framework introduced in Section 4.1 of the main paper is further illustrated in Figure~\ref{fig:supp_anno_demo}. At each iteration, the LLM (Claude Sonnet 3.5v2) proposes caption candidates, receives similarity-based feedback via IR-CLIP, and produces improved captions, ultimately converging to a final description selected by a stronger LLM (Claude Opus 4).

\smallskip
\noindent \textbf{Further Discussion.} While closely related to recent captioning method proposed in~\cite{ashutosh_llms_2025}, our approach departs in two key ways. First, we forgo the fixed-prompt bootstrapping and aggressive truncation strategies (\eg, seeding $\sim$30K prompts and retaining only the top-50 generations per step). Such strategies implicitly assume strong priors about the underlying image distribution---assumptions that may hold for curated RGB datasets but could break down for unlabeled, heterogeneous IR imagery. Second, rather than discarding low-scored candidates, we retain them as explicit hard negatives. Leveraging the longer context available in modern LLMs, our framework jointly conditions on both high- and low-quality captions, using low scores as counterfactual signals of what is not present in the IR image.

\subsection{Degradations}

To simulate real-world image degradations, we generated three types of altered inputs: blur, darkness, and fog. Blur was introduced by applying Gaussian smoothing with radii $\{0, 5, 10, 15, 20\}$, producing a controlled reduction of high-frequency detail. Darkness was simulated by scaling image brightness using factors $\{1.0, 0.45, 0.3, 0.2, 0.1\}$, where lower values correspond to reduced illumination. Fog effects were synthesized by blending the original image with a semi-transparent light-gray layer at intensities $\{0.0, 0.7, 0.85, 0.92, 0.97\}$, thereby decreasing contrast and diffusing edges. The selected parameter values for blur radius, brightness, and fog intensity were chosen based on qualitative visual inspection to ensure perceptually meaningful and progressively increasing degradation levels. Together, these degradations approximate common adverse conditions encountered in real outdoor environments.

\section{Implementation Details}\label{sec:supp_impl}

\method is implemented within the LLaVA~\cite{liu_improved_2024} VLM backbone, where the vision encoder as well as the base language model remain frozen. Only the LLM LoRA adapters, the projection module $P$, and the fusion weights are updated. We make use of the pretrained CLIP  ViT-$L/14$~\cite{radford_learning_2021, liu_improved_2024} as our frozen image encoder ($E$), to extract features from RGB and IR images resized to $336 \times 336$. With a patch size of $14 \times 14$, each image is thus represented by 576 tokens, each with an embedding size of 1024.

\end{document}